\documentclass[12pt]{article}

\usepackage [latin1]{inputenc}
\UseRawInputEncoding 
\usepackage{scicite}
\usepackage{multirow}
\usepackage{times}
\usepackage{lineno}
\usepackage{graphicx}
\usepackage{tabularx}
\usepackage{amsmath}
\usepackage{algorithm,algpseudocode}
\usepackage[font=small]{caption}

\newenvironment{sciabstract}{%
\begin{quote} \bf}
{\end{quote}}

\title{Embodied Design for Enhanced Flipper-Based Locomotion in Complex Terrains}

\author
{Nnamdi Chikere,$^{1}$ John McElroy,$^{2}$ Yasemin Ozkan-Aydin$^{1\ast}$\\
\\
\normalsize{$^{1}$Department of Electrical Engineering, University of Notre Dame,}\\
\normalsize{Notre Dame, Indiana, IN 46556, USA}\\
\normalsize{$^{2}$School
of Mechanical and Materials Engineering, University College Dublin,}\\
\normalsize{Belfield, Dublin 4, D04 V1W8, Ireland}\\
\\
\normalsize{$^\ast$To whom correspondence should be addressed; e-mail: yozkanay@nd.edu.}
}

\date{}

\usepackage{xr}
\makeatletter

\newcommand*{\addFileDependency}[1]{
\typeout{(#1)}
\@addtofilelist{#1}
\IfFileExists{#1}{}{\typeout{No file #1.}}
}\makeatother

\newcommand*{\myexternaldocument}[1]{%
\externaldocument{#1}%
\addFileDependency{#1.tex}%
\addFileDependency{#1.aux}%
}

\myexternaldocument{Supplementary}

\begin{document} 


\baselineskip24pt


\maketitle


\begin{sciabstract}
Robots are becoming increasingly essential for traversing complex environments such as disaster areas, extraterrestrial terrains, and marine environments. Yet, their potential is often limited by mobility and adaptability constraints. In nature, various animals have evolved finely tuned designs and anatomical features that enable efficient locomotion in diverse environments. Sea turtles, for instance, possess specialized flippers that facilitate both long-distance underwater travel and adept maneuvers across a range of coastal terrains. Building on the principles of embodied intelligence and drawing inspiration from sea turtle hatchings, this paper examines the critical interplay between a robot's physical form and its environmental interactions, focusing on how morphological traits and locomotive behaviors affect terrestrial navigation. We present a bio-inspired robotic system and study the impacts of flipper/body morphology and gait patterns on its terrestrial mobility across diverse terrains ranging from sand to rocks. Evaluating key performance metrics such as speed and cost of transport, our experimental results highlight adaptive designs as crucial for multi-terrain robotic mobility to achieve not only speed and efficiency but also the versatility needed to tackle the varied and complex terrains encountered in real-world applications.

\end{sciabstract}



\section*{Introduction}

Despite significant advancements in robotic locomotion, navigating diverse landscapes for tasks such as search and rescue in complex environments (e.g., sandy terrains, wet forests, and regolith-covered landscapes), as well as responding to mudslides and avalanches, remains a formidable challenge for robotic systems \cite{rafeeq_locomotion_2021,aguilar_review_2016}. While conventional wheeled and legged robots excel on solid ground, they often struggle on granular media such as sand, grains, or pebbles \cite{arvidson_spirit_2010} due to the non-uniform and deformable nature of the terrain \cite{goldman_colloquium_2014}. Moreover, factors like high resistance to penetration, instability, and limited load-bearing capacity of granular terrains can impede the mobility of these robots, leading to issues such as entrapment or slippage \cite{qian_walking_2012,yussof_methods_2010}. 


\begin{figure*}[t] 
    \centering
    \includegraphics[width=\textwidth]{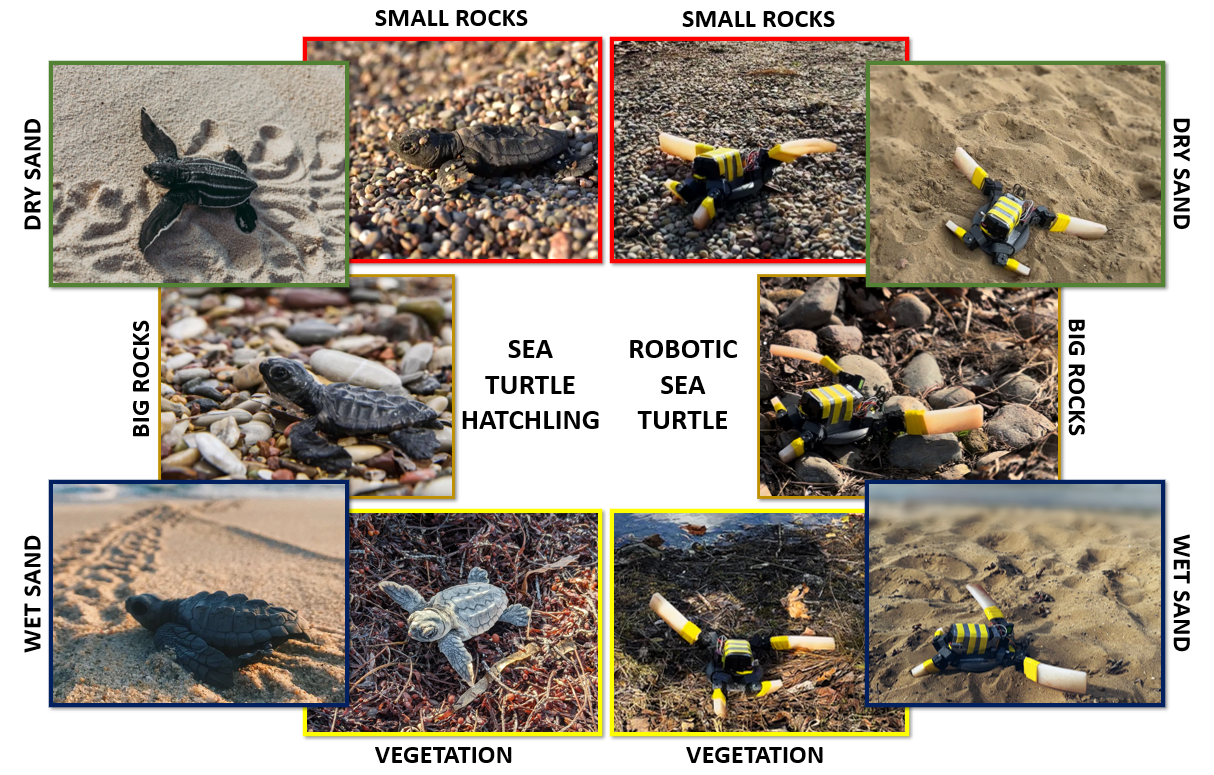}
    \caption{\textbf{Biological and robotic sea turtle hatchlings navigating diverse terrains:} Sea turtle hatchling (left) and its robotic counterpart (right) are shown traversing dry sand, small and big rocks, wet sand, and vegetation, illustrating the bio-inspired robot's design effectiveness and its capability to adapt to complex environmental conditions.}
    \label{fig:turtlevsrobot}
\end{figure*}

In addressing the limitations of traditional wheeled and legged robots, flipper-based locomotion offers a promising alternative. This concept draws inspiration from animals such as penguins, with their agile underwater propulsion using flippers \cite{hui_penguin_1988,hao_hydrodynamic_2023}, and seals, known for their maneuverability in both water and land \cite{leahy_role_2021}. Similarly, the fin-based locomotion of mudskippers, effective in terrestrial and aquatic settings, mirrors the adaptability of flipper-based systems, offering parallel insights for robotic design \cite{pace_mudskipper_2009,swanson_kinematics_2004}. Drawing inspiration from aquatic and amphibious animals, we can equip robots with flexible and powerful flippers, enhancing adaptable propulsion and maneuverability in diverse environments, from aquatic to granular terrains \cite{ijspeert_biorobotics_2014,melo_animal_2023,iida_biologically_2016,pfeifer_self-organization_2007,gravish_robotics-inspired_2018}.  

Among the various examples of flipper-based locomotion in nature, sea turtles are particularly notable for their adeptness in traversing multiple terrains starting from a very early stage in their lives (Fig. \ref{fig:turtlevsrobot} - left). These animals have evolved locomotion mechanics, allowing them to navigate efficiently in water - their primary habitat - and on land, where they venture mainly for reproduction, including nesting and hatching activities \cite{hirth_aspects_1980}. Their specialized body design and unique flippers allow them to dynamically adjust to varying terrain conditions, from aquatic environments to sandy beaches and rocky shorelines \cite{lutz_biology_2017}. This adaptability and proficiency in handling different granular media make sea turtles an ideal biological model for advancing robotic design. By studying and replicating the sea turtle's locomotion mechanics, our research aims to develop robotic systems capable of efficient navigation across various terrestrial environments (Fig. \ref{fig:turtlevsrobot} - right), enhancing bio-inspired robots' versatility and practical applicability.

\begin{figure*}[t] 
    \centering
    \includegraphics[width=\textwidth]{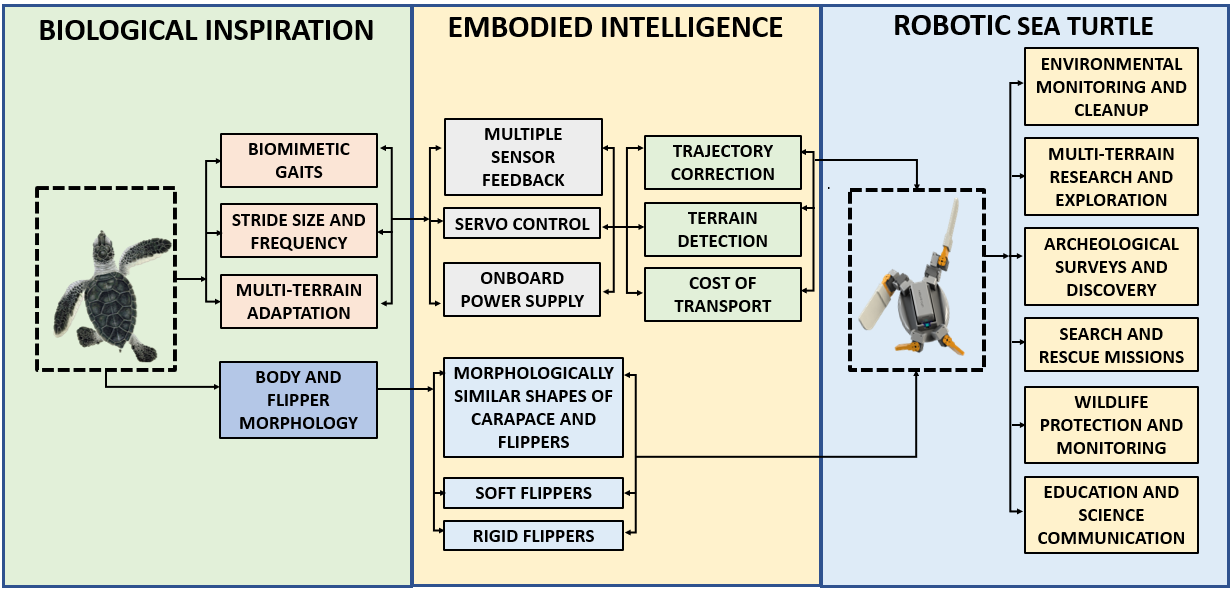}

    \caption{\textbf{Conceptual framework of the robotic sea turtle:} The integrated approach combining biological inspiration with advanced robotics. The leftmost block details the biological aspects that influence the design. The center block describes the robot's embodied intelligence features, including sensor feedback, servo control, and power management. The rightmost block showcases the diverse applications of the robotic sea turtle.}
    \label{fig:summary}
\end{figure*}

Extensive research in bio-inspired robotics has explored various aspects of sea turtle locomotion. Biologists have focused on the distinct characteristics and biomechanical principles that facilitate their effective movement \cite{lutz_biology_2017,avens_responses_2003,renous_comparison_1993}. Concurrently, researchers in robotics have been examining techniques to replicate these characteristics in robotic designs. Recent studies have explored various aspects, from the mechanics of sea turtle-inspired robots \cite{mazouchova_flipper-driven_2013} to exploring the impact of different gaits of locomotion in sea turtle robots \cite{han_mechanism_2011,song_turtle_2016,wu_fully_2022}. Mazouchova \cite{mazouchova_flipper-driven_2013} found that a sea turtle-inspired robot's terrestrial movement is influenced by flipper penetration and stroke size. Zhang \cite{zhang_amphihex-i_2016} and Zhong \cite{zhong_locomotion_2016} both explored the locomotion performance of amphibious robots with transformable flipper legs, with Zhang focusing on muddy terrain and Zhong on various terrains and underwater. Both studies highlighted the importance of flipper leg rotation speed and stiffness in achieving stable locomotion. The influence of the stiffness of the flipper legs on the locomotion performance \cite{zhong_locomotion_2016} as well as the load-bearing capacity associated with each locomotive gait \cite{jansen_bio-inspired_2017} has also been investigated. More recently, Baines et al. \cite{baines_multi-environment_2022} took inspiration from chelonian environmental adaptations to develop a morphing amphibious robotic limb that can switch between a streamlined flipper and a load-bearing leg, enhancing performance in both aquatic and terrestrial environments.
\begin{figure*}[!t] 
    \centering
    \includegraphics[width=\textwidth]{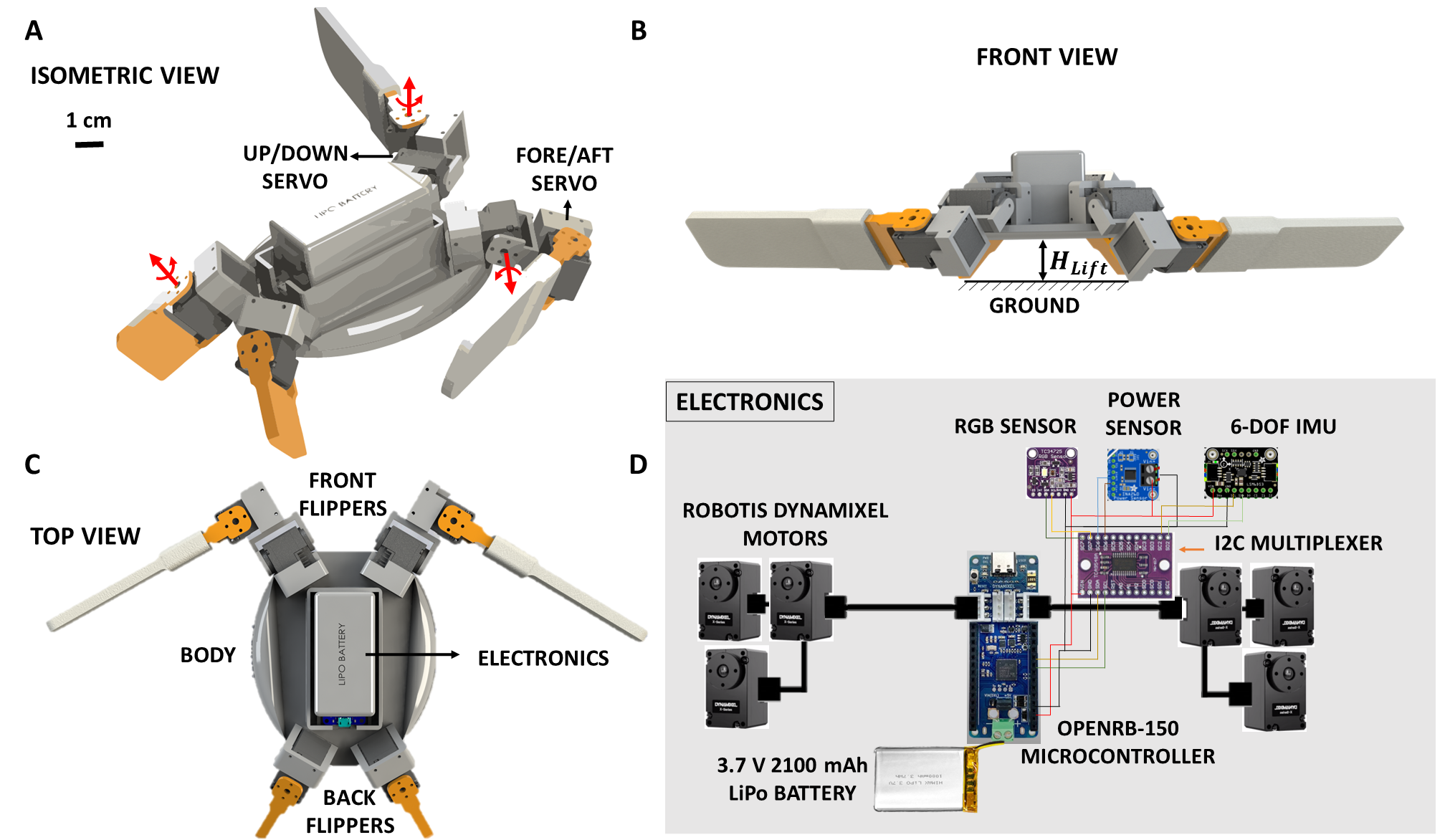}
    \caption{\textbf{Anatomical illustration of the robotic sea turtle:} (\textbf{A}) Isometric CAD view, highlighting the rotation axes of the servos for flipper actuation. (\textbf{B}) Front view, showing the lift height of the robot above the ground. (\textbf{C}) Top view, showing the robot's symmetry and flipper arrangement. (\textbf{D}) Electronic components}
    \label{fig:fullimage}
\end{figure*}
Findings from previous works on sea turtle-inspired robotics for real-world applications emphasize the need for ongoing research to refine their design and functionality. Despite current advancements, there remain notable gaps in the research, particularly concerning the relationship between gait patterns and flipper/body morphology across various challenging terrestrial terrains and the energy efficiency of these systems. Previous works exhibited some limitations, including the exclusive use of front flippers \cite{mazouchova_flipper-driven_2013,luck_lab_2017}, reliance on a single terrestrial gait\cite{baines_multi-environment_2022}, notable gaps in exploring the relationship between gait patterns and flipper/body morphology for various challenging terrestrial terrains, and an analysis of the performance of the robot in only one type of terrestrial terrain. These constraints require a more holistic approach that considers the interplay between gait patterns, flipper morphology, and their implications on robotic design in various environments.


In our study, we focus on the impact of flipper/body morphology and gait patterns on the terrain adaptability of a flippered robot inspired by the morphology and locomotion mechanics of sea turtle hatchlings. A simplified robotic system (Fig. \ref{fig:fullimage}) was developed, tailored to function efficiently in diverse environments, specifically sand and rocky terrains. This system integrates a combination of gaits and different flipper stiffness and numbers, allowing for performance assessment in terms of average displacement and cost of transport. Our experimentation demonstrates that efficient locomotion requires distinct gait and flipper stiffness adaptations in different environments. 

The findings from our study provide crucial insights into the design and control mechanisms for robotic systems aimed at emulating the locomotion of sea turtles. The unique features of our robot include its trajectory correction abilities, terrain recognition, and adaptive gait adjustments, which are coupled with the influence of morphology on its steering ability. The robot is designed to be low-cost, centimeter-scaled, robust, and autonomous, demonstrating versatile capabilities in navigation and interaction with various environments. Fig. \ref{fig:summary} offers a comprehensive overview of the conceptual framework that drives the design and application of our bio-inspired robotic sea turtle. The robot represents a significant advancement in robotics, embodying the principles of bio-inspired design and embodied intelligence to achieve deformable media locomotion as well as terrain adaptiveness. This innovative approach enables the robot to navigate through diverse environments with the efficiency of its natural counterpart, highlighting its practical applicability in real-world scenarios such as environmental monitoring, search and rescue operations, and marine biology research. By integrating sensors, actuators, and control algorithms, the robot autonomously adapts to changing conditions, optimizing its movement through gait changes. This demonstrates the feasibility of replicating complex biological movements in robotic systems and opens new avenues for their use in scientific exploration, conservation efforts, and educational initiatives, showcasing the transformative potential of bio-inspired robotics in addressing global challenges.

\begin{figure*}[!h] 
    \centering
    \includegraphics[width=\textwidth]{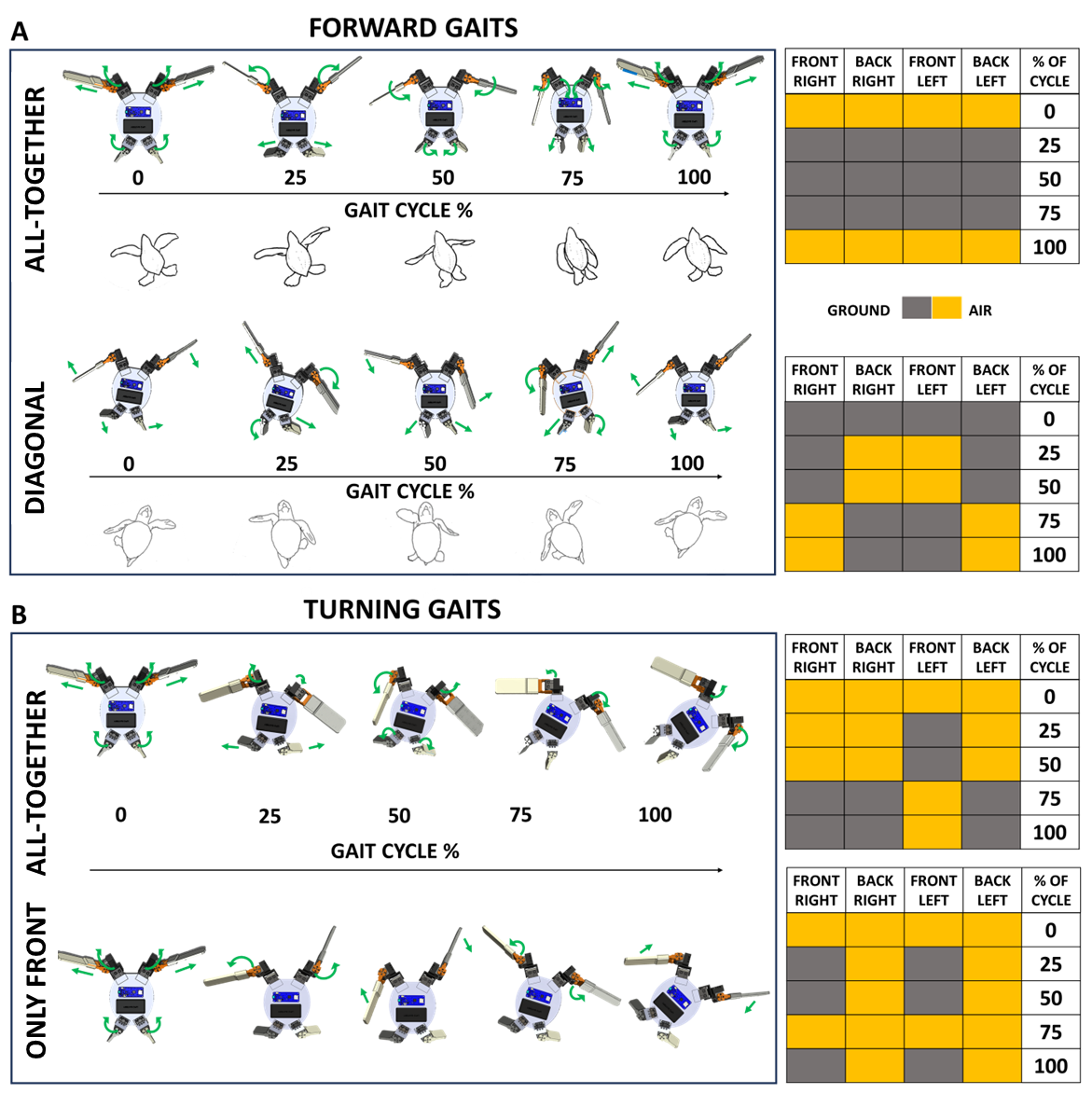}
    \caption{\textbf{Illustration of gait patterns:} (\textbf{A}) Forward Gaits displaying 'All-together' (top row) and 'Diagonal' (bottom row) patterns with corresponding gait cycle diagrams. (\textbf{B}) Turning Gaits exhibiting 'All Flippers Together' (top row) and 'Only Front Flippers' (bottom row) patterns alongside their respective gait cycle diagrams. Each diagram visualizes the flipper contact with the ground (gray) and aerial phase (yellow) throughout the gait cycle. The green arrows indicate the directional movement of the flippers, emphasizing the active propulsion phases of forward and rotational movement.}
    \label{fig:gaitsequence}
\end{figure*}

\section*{Results}
\subsection*{Biological Inspiration and Robot Design}

Drawing inspiration from the sea turtle hatchlings' morphology (Fig. \ref{fig:turtlevsrobot}(left)), our robot mimics their broad surface area and an upturned plastron to maintain traction and prevent sinking on soft or irregular ground. The robot's design emulates its oval body, and we incorporate interchangeable flippers (soft and rigid) for efficient terrain navigation and to investigate the impact of flipper stiffness. Additionally, the turtle's "crutching" motion is replicated to enhance stability and mobility on challenging substrates. 

Studies have shown that sea turtles commonly exhibit two distinct gait patterns: the 'Cheloniid' diagonal gait and the 'Dermochelyid' all-together gait \cite{lutz_biology_2017,wyneken_sea_1996}. The 'Cheloniid' pattern, characteristic of species like the loggerhead (\textit{Caretta caretta}), involves hatchlings using diagonally opposite limbs for crawling. This movement creates thrust through specific parts of their flippers and is a typical pattern across all Cheloniid hatchlings. On the other hand, the 'Dermochelyid' pattern observed in leatherback turtles (\textit{Dermochelys coriacea}) consists of a "swing and stance" limb cycle, where all four flippers move together. In our work, we implement both gaits and evaluate their impact on the performance of the robot.

\subsection*{Gait Implementation}
We implement two biological gait patterns on our robot, i.e., the all-together gait (Fig. \ref{fig:gaitsequence}A - top row) and the diagonal gait (Fig. \ref{fig:gaitsequence}A - bottom row). The gait sequence algorithms that imitate the motion patterns of the flippers during locomotion are shown in movie S2. The all-together gait is characterized by a simultaneous downward thrust and upward lift of all flippers. On the other hand, the diagonal gait, involving staggered actuation of diagonally opposite flippers, creates a dynamic tilt in the robot's body, ranging from a high point at the rear to a low point at the front. This results in an alternating tilt angle of roughly $20^\circ$ between diagonally opposite flippers. These gaits also differ in the sequence and extent of actuation of the robotic flippers during a locomotion cycle.

The locomotion cycle and sequence are mainly governed by the actuation of the 2-degree-of-freedom (DoF) shoulder joint of the front flippers (Fig. \ref{fig:fullimage}A), which comprises two motors linked via a mechanism that permits the motion of the end effector, flipper, in two planes. The base motor that is connected to the body rotating within an angular range of $\alpha = -55^\circ$ (downward) to $90^\circ$ (upward) in the z-plane controls the up/down rotation of the fore/aft motor (Fig. S1B). Subsequently, the fore/aft motor governs the swing of the fore flippers in the x-plane within an angular limit of $\beta = -10^\circ$ (backward) to $75^\circ$ (forward) from their neutral positions (Fig. S1B). Fig. S2 shows a plot of the trajectory of the servomotor's angle positions during the gaits, highlighting the synchronized movement of servos during the all-together gait and the alternating servo positions during the diagonal gait, detailed over two complete cycles. During the full upward extension of the flipper, this mechanism lifts the flipper's tip to a height of about 9.2 cm off the ground. When the flippers push down maximally during all-together gait, the robot's body ($H_{Lift}$) is approximately 4 cm off the ground. The 1-DoF back flippers, responsible for directional control, rotate horizontally from $\gamma = 90^\circ$ to $-30^\circ$.

\subsection*{Experimental Setup and Results}

Experiments to investigate the locomotion mechanics were conducted on multiple terrains, including dry sand, rocky terrain (pebbles), wet sand, flat foam surface, foam stairs, and sandy inclines, as depicted in Fig. S\ref{fig:terrains}A. During testing, a 160x100 $cm^2$-bordered area for each terrain type was established to contain the robot's movement. Four main flipper configurations were examined for each gait (Fig. S\ref{fig:terrains})B - the robot was equipped with either soft or rigid flippers on both ends or just on the front ends with no rear flippers. 

\begin{figure*}[t] 
    \centering
    \includegraphics[width=\textwidth]{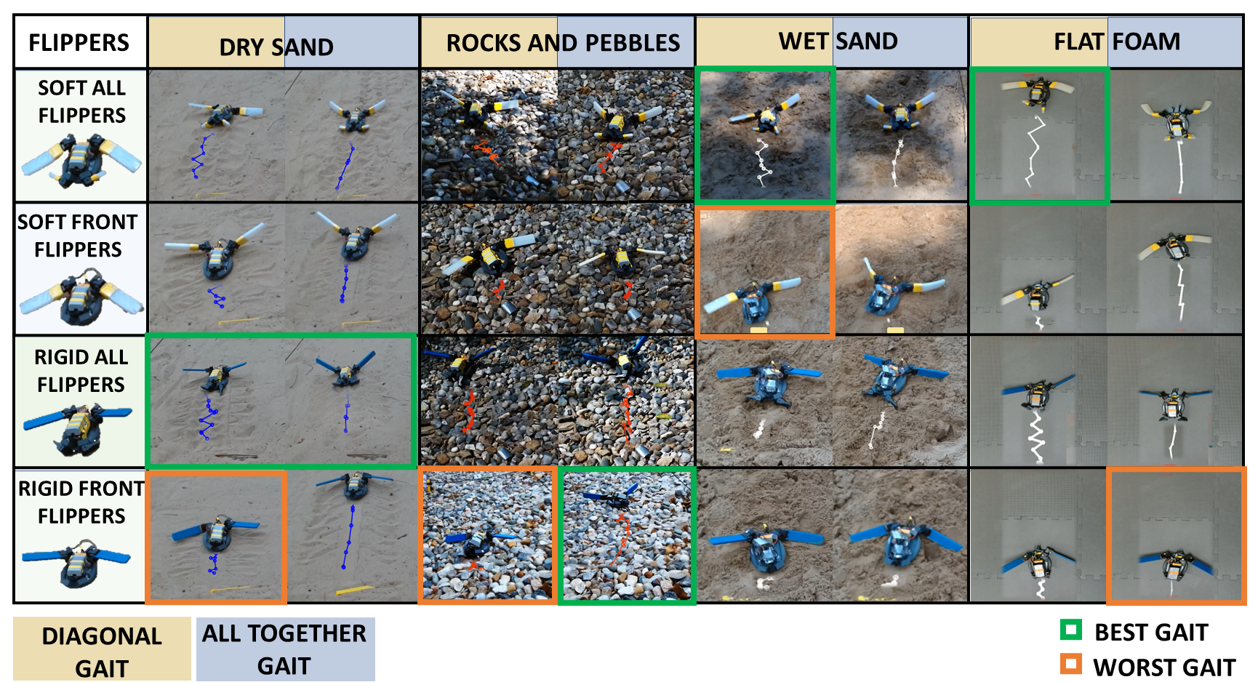}
    \caption{\textbf{Gait efficiency across varied terrains:} The robot's performance with different flipper types (soft all flippers, soft front flippers, rigid all flippers, and rigid front flippers) and gait patterns (diagonal and all-together) across four terrain types: dry sand, rocks and pebbles, wet sand, and flat foam. The best and worst gaits for each flipper and terrain combination are highlighted with green and orange borders, respectively. The trajectory tracking of the robot's Center of Mass (CoM) during locomotion is shown in the blue, orange, and white lines}
    \label{fig:terrainplots}
\end{figure*}

The directional control was guided using the Inertial Measurement Unit (IMU) to ensure a consistent trajectory during locomotion, and the robot's power consumption was measured for each experimental run. Two web cameras (Logitech C920x HD Pro, 30 frames per second) were installed, one for capturing the top/back view and another for recording a side view of the robot's locomotion dynamics. The robot was positioned near one end of the border and powered on. We conducted three trials for each experiment and recorded the average displacement in terms of body length per cycle (BL/cycle) and power consumption. The best and worst performing configurations for each terrain are as shown in Fig. \ref{fig:terrainplots}.

\subsubsection*{Dry Sand Tests}
    In dry sandy environments, our experiments (movie S3) compare different flipper configurations and gaits, and our results (Fig. \ref{fig:terrainperformance}A) reveal that using all flippers - whether soft or rigid — generally ensures effective locomotion. Notably, the rigid flipper setup using all-together and diagonal gaits stands out, yielding the highest measured displacement of $0.83 \pm 0.03$ BL/cycle. Additionally, the most energy-efficient movement, as indicated by the lowest Cost of Transport (CoT) of \(9.04\), was achieved with the rigid flippers in a diagonal gait. The soft flipper configuration was also effective, with displacements of $0.78 \pm 0.07$ BL/cycle for the all-together gait and \(0.74\) BL/cycle for the diagonal gait. The front-only flipper configurations recorded lower displacements ranging from 0.29 $\pm$ 0.04  to 0.51 $\pm$ 0.05 BL/cycle with the exception of the rigid front flippers with the all-together gait, which achieved a competitive displacement of $0.81 \pm 0.03$ BL/cycle.
    These results indicate that while the use of all four flippers consistently enhances locomotion effectiveness on dry sand, there is potential for optimizing the front flipper in an all-together gait in certain scenarios or specific robot designs to achieve good performance on sandy terrain.

\subsubsection*{Rocky Terrain Tests}

\begin{figure*}[!t] 
    \centering
    \includegraphics[width=\textwidth]{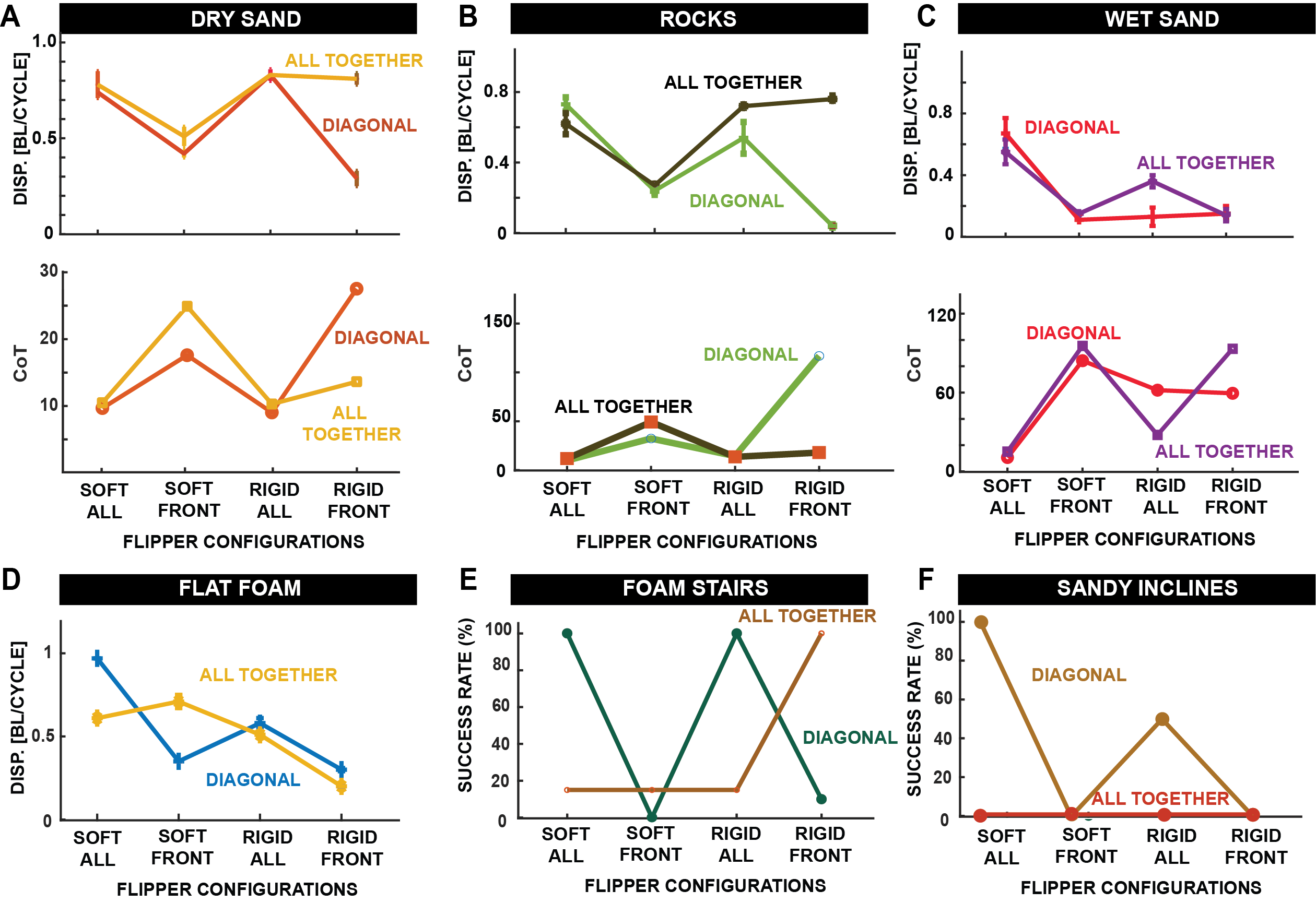}
    \caption{\textbf{Performance metrics on various terrains:} The robot's CoM displacement per cycle and Cost of Transport (CoT) across different terrains with each flipper configuration (soft all, soft front, rigid, all, and rigid front) and gait patterns (diagonal and all together). (\textbf{A}) Dry sand, (\textbf{B}) Rocks, (\textbf{C}) Wet sand. (\textbf{D}) Flat foam and (\textbf{E}) Foam stairs, (\textbf{F}) Sandy inclines display success rates, illustrating the effectiveness of gait patterns in overcoming terrain obstacles. }
    \label{fig:terrainperformance}
\end{figure*}

    Rocky terrains, with their inherent irregularities as depicted in (Fig. S\ref{fig:terrains}A, movie S4), pose a significant challenge, demanding superior grip and stability for successful navigation. The terrain consisted of pebbles and rocks with sizes ranging from $3.5$ to $6$ cm. The rigid front flippers in an all-together gait outperformed the rest with a displacement of 0.76 $\pm$ 0.02 BL/cycle (Fig. \ref{fig:terrainperformance}B). This likely stems from the gait pattern, which permits the robot to glide over the rocks, leveraging the strong support provided by the rigid front flippers. However, from an efficiency perspective, the soft flippers in a diagonal gait achieved the lowest CoT of \(9.84\) while also closely competing in displacement, recording an average displacement of \(0.73 \pm 0.04\) BL/cycle. This efficiency might be due to the soft material's adaptability to the terrain's unevenness, ensuring consistent ground contact, while the diagonal gait offers balance. However, aside from the rigid front-only all-together gait, which exhibited a remarkable performance, other front-only flipper configurations underperformed in rocky terrain tests. This exception suggests that design optimizations that concentrate on the functionality of the front flippers could lead to enhanced locomotion in rocky environments.

\subsubsection*{Wet Sand Tests}

    The wet sand terrain (Fig. S\ref{fig:terrains}A, movie S5), reflective of real-world challenges found in cohesive granular materials like beach sand, posed a significant challenge in our tests. Surface tension makes wet sand cohesive \cite{mitarai_wet_2006}, and our results (Fig. \ref{fig:terrainperformance}C) emphasized the superiority of flipper flexibility in such conditions. Soft flippers in a diagonal gait achieved the highest displacement at \(0.67 \pm 0.10\) body lengths (BL) per cycle and the most efficient Cost of Transport (CoT) of $10.61$. This efficacy is attributed to the soft flippers' ability to contour to the bumpy terrain, enhancing traction, while the diagonal gait provides consistent propulsion. The all-together gait with soft flippers also showed favorable outcomes with a displacement of \(0.55 \pm 0.08\) BL/cycle. Conversely, rigid flippers fell behind in performance, recording a displacement of \(0.13 \pm 0.06\) BL/cycle with the diagonal gait and \(0.36 \pm 0.04\) BL/cycle with the all-together gait, highlighting that flipper stiffness is a detriment in moist, non-uniform terrains. Furthermore, experiments demonstrated that setups with only front flippers were less effective on wet sand for both gait patterns tested. Specifically, the configuration with soft front flippers resulted in displacements of \(0.11 \pm 0.00\) BL/cycle and \(0.15 \pm 0.00\) BL/cycle for the diagonal and all-together gaits, respectively. Similarly, the rigid front-only flipper setup recorded displacements of \(0.15 \pm 0.04\) BL/cycle for the diagonal gait and \(0.14 \pm 0.04\) BL/cycle for the all-together gait. These findings highlight the substantial impact of flipper configuration on mobility in challenging environments.

\subsubsection*{Flat Foam Tests}

    Our experiment setup on the flat foam surface consisted of a $100 cm$ x $100 cm$ square foam block (Fig. S\ref{fig:terrains}A, movie S6). As shown in Fig. \ref{fig:terrainperformance}D, our results revealed varied performances for different flipper and gait combinations. The homogeneous and predictable nature of the flat foam surface allowed for a clear assessment of the efficiency and efficacy of each setup. The soft flippers in a diagonal gait demonstrated the highest displacement, achieving $0.97 \pm 0.00$ BL/cycle, with the all-together gait also performing well. Notably, configurations using only front flippers showed significant differences; the soft front flippers in an all-together gait outperformed their diagonal gait counterpart with recorded displacements of $0.71 \pm 0.04$ BL/cycle and $0.35 \pm 0.01$ BL/cycle, respectively. Rigid flippers had mixed results, with the diagonal gait yielding better displacement than the all-together gait but still falling short compared to soft flippers. The rigid front-only flippers significantly showed limited movement on this terrain, particularly in the all-together gait configuration with just $0.20 \pm 0.00$ BL/cycle. These outcomes underscore the importance of flipper flexibility and movement gaits in optimizing locomotion across flat, hard surfaces.

\subsubsection*{Foam Stairs Tests}

The experimental trials on a staircase composed of five 2.5 cm high foam blocks (Fig. S\ref{fig:terrains}A, movie S7) introduced a unique set of challenges, primarily due to the vertical ascent and the climb required for successful navigation. With the diagonal gait, the soft flipper configuration consistently ascended and descended the entire staircase, averaging 17 \(\pm\) 3.5 cycles to complete the path and a 100\% success rate. However, with the all-together gait, its ascent efficiency was reduced to a 15\% success rate due to complications arising from the interaction of the back flippers with the stairs, though descent remained unaffected with a 100\% success rate. The front-only soft flippers could not climb the stairs in either gait and with the all-together gait, it achieved a 15\% ascent success rate by climbing a single stair. The rigid full flipper setup mirrored the soft flipper's performance in the diagonal gait, completing both ascent and descent with a 100\% success rate. However, the ascent success rate dropped to 15\% with the all-together gait due to back flipper interference, but the descent remained successful.
The front-only rigid flippers performed poorly on the stairs with either gait, achieving 10\% success rate in ascending partially up a single step, although their descent was successful. In summary, full flipper setups with diagonal gait performed the best (100\% success rate), and front-only configurations performed poorly, as shown in the results plot in Fig. \ref{fig:terrainperformance}E.

\subsubsection*{Sandy Inclines Tests}

The sandy inclines test setup (Fig. S\ref{fig:terrains}A, movie S8) consisted of two 20-degree angled slopes with a valley between them, simulating a real-life, challenging terrain. The soft flipper with the diagonal gait successfully climbed both the hilly slopes and emerged from the valley, thus showing a 100\% success rate. The success of the soft flippers with diagonal gait illustrates the configuration's capability to adapt and maintain traction on inclines. On the other hand, the soft flipper with the all-together gait failed to ascend the slopes, as the robot tended to burrow into the sand instead of ascending, indicating a need for gait modification in such conditions.
Regardless of gait type, the front-only soft flipper configurations could not climb the slopes, displaying a clear shortcoming in this particular terrain with a 0\% success rate. This highlights the critical need for rear flipper support in uphill tasks. As regards the full rigid flipper configuration, the diagonal gait allowed the robot to ascend the first slope successfully but not climb out of the valley, resulting in a 50\% success rate. Similar to the soft flipper configuration, with the all-together gait, the robot could not climb the slopes. Finally, regardless of gait type, the front-only rigid flipper configurations failed to ascend the slopes, presenting a 0\% success rate. These results (Fig. \ref{fig:terrainperformance}F) show that the diagonal gait paired with a full soft flipper setup is best suited for sandy inclines, ensuring a complete success rate. In contrast, front-only setups failed across the board, underscoring the contribution of the rear flippers in overcoming steep, granular terrains.


\subsubsection*{Turning Gait}

Turning efficiency is crucial for our robotic system's adaptability across diverse terrains. Our experiments explored the impact of flipper configurations and stiffness on the robot's ability to execute turns with precision and speed. The turning gaits experiments aimed to identify optimal strategies for maneuvering in environments that require agile and responsive movement to achieve a change in direction.

\begin{figure*}[!t] 
    \centering
    \includegraphics[width=\textwidth]{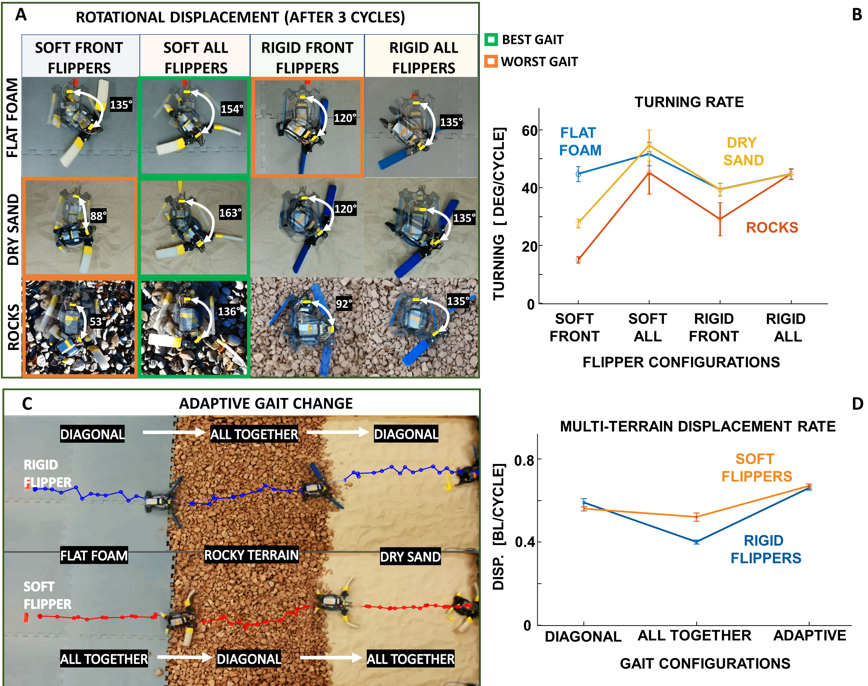}
      \caption{\textbf{(A) Rotational displacement on various terrains:} The robot's rotational displacements with different flipper configurations across flat foam, dry sand, and rocks. The angles indicate degrees turned after three gait cycles, with overlays showing the best (green) and worst (red) performing gaits. \textbf{(B) Comparison of gait efficiency :} The turning rate in degrees per cycle for each flipper configuration. \textbf{(C) Gait adaptation:} The robot's path across a segmented terrain setup, with indications of gait changes (Diagonal Gait, All Together Gait) corresponding to the type of flippers (Rigid, Soft) and the terrain encountered (Flat Foam, Rocky Terrain, Dry Sand). The Center of Mass (COM) trajectory tracking is shown in red and blue lines for the rigid and soft flippers. \textbf{(D) Multi-terrain displacement rate:} Comparison of the displacement rate in body lengths per cycle for different gait configurations (Diagonal, All Together, Adaptive) and flipper types (Soft, Rigid).}
    \label{fig:turnandgtransit}
\end{figure*}

Our experiments investigated two different gait strategies, the "Front Flippers" and "All Flippers," on three terrains: flat foam, rocky terrain, and dry sand (see Fig. \ref{fig:turnandgtransit}A, movie S9). The 'Front Flippers' strategy involves using only the two front flippers to initiate rotation — with one flipper pushing forward while the other pulls back, not engaging the back flippers. In contrast, the 'All Flippers' strategy employs all four flippers, with front flippers working in an alternating pattern while the back flippers support the motion, creating a dynamic balance that aids in turning.

We quantified the robot's turning efficacy by measuring the rate of rotational displacement needed to complete a full 360-degree turn. Our findings (Fig. \ref{fig:turnandgtransit}B), expressed in degrees per cycle (deg/cycle), highlight the impact of flipper material and configuration on rotational capabilities.

\paragraph{Flat Foam Surface Turning:}
The data reveals that on a flat foam surface, the 'Soft Flipper - All Flippers' configuration outperformed the rest, achieving the highest average rotational speed of 51.47 \(\pm\) 3.74 deg/cycle. This suggests that the flexibility of soft flippers combined with the coordinated movement of all flippers contributes significantly to turning agility. In contrast, configurations employing only front flippers, whether soft or rigid, yielded lower turning rates, with the 'Soft Flipper - Front Flippers' at 45.02 \(\pm\) 2.39 deg/cycle and the 'Rigid Flipper - Front Flippers' at 40.02 \(\pm\) 1.99 deg/cycle. Interestingly, the 'Rigid Flipper - All Flippers' setup showed a turning rate closely comparable to the 'Soft Flipper - Front Flippers,' with an average of 45.01 \(\pm\) 1.61 deg/cycle, underscoring that the complete engagement of all flippers can somewhat compensate for the lack of flexibility in rigid flipper.
These results guide us toward configurations that maximize the robot's turning efficiency, favoring setups that engage all flippers, particularly with soft materials, for enhanced maneuverability on flat terrains.

\paragraph{Rocky Terrain Turning:}
On rocky terrain, turning efficiency varied significantly with flipper type and gait configuration. The 'Soft Flipper - All Flippers' setup showed a high average turning rate of 45.41 \(\pm\) 6.81 deg/cycle, indicating that the soft material's adaptability is beneficial even amidst irregular surfaces. However, the high standard deviation suggests less predictability in performance, possibly due to the variable nature of rocky terrain. The 'Soft Flipper - Front Flippers' configuration yielded a lower turning rate of 17.45 \(\pm\) 1.03 deg/cycle, highlighting the limitations of using front flippers alone in complex terrains. In contrast, the 'Rigid Flipper - All Flippers' setup demonstrated comparable turning efficiency to the soft all-flipper setup, with an average rate of 45.01 \(\pm\) 1.72 deg/cycle. The lower standard deviation here reflects a more consistent performance, likely due to the rigidity providing stable pivot points on uneven ground. Notably, the 'Rigid Flipper - Front Flippers' achieved an average of 30.52 \(\pm\) 5.34 deg/cycle, underscoring that while rigidity may provide stable turning points, the lack of rear flipper actuation limits overall turning capability. 
These results from rocky terrains suggest that while all-flipper configurations generally offer improved turning capabilities, the advantages of soft versus rigid flippers may vary depending on the uniformity and predictability of the surface. The findings emphasize the importance of considering flipper material and gait configuration when designing robots for maneuverability in complex environments.

\paragraph{Dry Sand Turning:}Our turning rate analysis on dry sand terrain reveals some interesting insights. The 'Soft Flipper - All Flippers' configuration showed a superior average turning rate of 54.23\(\pm\)4.98 degrees per cycle, suggesting that the inherent flexibility of soft flippers offers substantial advantages in granular media, likely due to the increased surface area contact and the flippers' ability to scoop and push against the loose sand. Meanwhile, 'Soft Flipper - Front Flippers' demonstrated a lower average turning rate of 29.22 \(\pm\) 1.35 degrees per cycle. Compared to the all-flipper setup, this reduced efficiency reinforces that rear-flipper engagement is critical for optimizing maneuverability in sandy conditions. The 'Rigid Flipper - Front Flippers' achieved an average rate of 40.01 \(\pm\) 1.91 degrees per cycle, indicating that rigidity can somewhat compensate for the lack of rear flipper action by providing firm leverage points in the sand. However, when all rigid flippers were engaged, the configuration maintained a consistent turning rate of 45.00 degrees per cycle, highlighting the predictability and reliability of rigid flippers in dry sand when a coordinated gait is used.
These results from dry sand conditions underscore that while all-flipper configurations generally provide better turning rates, the choice between soft and rigid flippers should be tailored to the specific demands of the terrain, balancing flexibility and stability for optimal performance.

In summary, our turning experiments suggest that flipper stiffness and the number of flippers engaged significantly impact the robot's ability to maneuver efficiently. Soft flippers, particularly when all actuated, provide high degrees of rotation per cycle across various terrains, with their flexibility allowing for greater surface area contact and adaptability to the terrain's inconsistencies. Rigid flippers, while offering less adaptability, provide a consistent and predictable turning rate, especially when all are engaged, which may be beneficial in environments where precision is valued over a higher rotational displacement rate. The optimum configuration (soft or hard, front only or all flippers), therefore, depends on the particular terrain challenges and the maneuvering requirements of the task.

\subsubsection*{Trajectory Correction}
To enhance our robot's navigational accuracy, we integrated a trajectory correction algorithm that relies on data from an Inertial Measurement Unit (IMU). This system continuously monitors the robot's orientation and acceleration to detect deviations from the set path. By processing this real-time information, the algorithm can make dynamic adjustments to the robot's movements, correcting its course to stay aligned with the planned trajectory. This feature is crucial for conducting precise experiments and maintaining effective locomotion across different terrains, where unanticipated obstacles or surface irregularities might otherwise lead to off-course travel.
The trajectory correction feature functions as in Algorithm \ref{algo1}: \\

\begin{algorithm}
\caption{Trajectory Correction}
\label{algo1}
\begin{algorithmic}[1]
\State Initialize the Inertial Measurement Unit (IMU) on the robot.
\State Set \texttt{target\_trajectory} to the desired straight path.
\State Begin loop.
\Procedure{Trajectory Correction}{IMU}
	\State \texttt{current\_trajectory} $\gets$ \Call{Read\_IMU\_Data}{IMU}
	\State \texttt{deviation} $\gets$ \Call{Calculate\_Deviation}{current\_trajectory, target\_trajectory}
	\If{\texttt{deviation} $> 15$ cm}
		\If{\texttt{deviation} is leftward}
			\State \texttt{correction\_gait} $\gets$ ``Right Correction Gait''
		\ElsIf{\texttt{deviation} is rightward}
			\State \texttt{correction\_gait} $\gets$ ``Left Correction Gait''
		\EndIf
		\State \Call{Execute\_Correction}{correction\_gait}
	\Else
		\State \Call{Execute\_Movement}{``Straight Gait''}
	\EndIf
	\State \textbf{return} to step 5 for continuous correction.
\EndProcedure 
\end{algorithmic}
\end{algorithm}

In the algorithm, the robot continuously reads data from the IMU to assess its current trajectory against the intended path. If the deviation exceeds 15 cm, the robot determines the direction of the deviation and executes a correction gait to adjust its course. This corrective action involves actuating the flippers to counteract the deviation—employing a 'Right Correction Gait' for leftward deviations and a 'Left Correction Gait' for rightward deviations. If the robot is on course, it continues with a 'Straight Gait.' The process repeats in a loop to maintain the correct trajectory continuously.

\subsubsection*{Gait Transition Experiments}

Building on the insights from our locomotion tests, we equipped our robotic system with a feature that allows for gait transition, optimizing its movement to correspond with the detected terrain. This adaptive gait mechanism functions by a terrain recognition system that utilizes a color sensor, enabling the robot to identify different surfaces and adjust its gait accordingly.

\begin{algorithm}
\caption{Sea Turtle Robot Terrain Adaptation}
\label{algo2}
\begin{algorithmic}[1]
\State Initialize the color sensor on the robot.
\State Begin loop.
\Procedure{Adaptive Gait}{color\_sensor}
    \State color $\gets$ \Call{Detect\_Color}{color\_sensor}
    \If{color is grey}
        \State terrain $\gets$ ``Hard Ground''
    \ElsIf{color is red}
        \State terrain $\gets$ ``Rocks''
    \ElsIf{color is light brown}
        \State terrain $\gets$ ``Sand''
    \Else
        \State terrain $\gets$ ``Unknown''
    \EndIf
    \State gait $\gets$ \Call{Select\_Gait}{terrain}
    \State \Call{Execute\_Movement}{gait}
    \State \textbf{return} to step 4 for continuous adaptation.
\EndProcedure
\end{algorithmic}
\end{algorithm}
For practical application, we conducted a series of experiments using both rigid and soft flippers (see Fig. \ref{fig:turnandgtransit}C, movie S10). The algorithm we developed processes input from the color sensor to identify the type of terrain. Upon terrain recognition, the algorithm selects the gait experiments shown to be most efficient for that specific surface—whether it be the diagonal gait for soft, uneven surfaces or an all-together gait for stable, flat terrains.
Algorithm \ref{algo2} summarises the terrain recognition and gait adaptation system.

Experiments were conducted within a confined area measuring 105 cm across, featuring segments of three different terrains: flat foam, rocks, and sand, each designated by a specific color—grey for flat foam, red for rocks, and light brown for sand. The robot's objective was to navigate this multi-terrain box, adapting its gait dynamically in response to the color sensed by its onboard sensor. Concurrently, the Inertial Measurement Unit (IMU) was utilized for trajectory correction to maintain precise navigation.

We tested the robot in both the rigid and soft flipper configurations to measure its displacement rate per cycle, focusing on the impact of the adaptive gait on performance across varying terrains compared to its performance observed when using a constant gait. As shown in Fig. \ref{fig:turnandgtransit}C, the adaptive gait mechanism allowed smooth transitions between gaits, enhancing the robot's locomotive efficiency. The trajectory followed by the robot, equipped with both rigid and soft flippers, is shown in Fig. \ref{fig:turnandgtransit}C, highlighting where gait transitions occurred to accommodate different terrains.

The measured displacement rate data for both the adaptive gait changes and the single gait experiments are presented in Fig. \ref{fig:turnandgtransit}D. The data reveals that for soft flippers, the 'Adaptive Gait Change' achieved the highest displacement at \(0.66 \pm 0.01\) BL/cycle, surpassing both the 'Diagonal Gait' at \(0.59 \pm 0.01\) and the 'All Together Gait' at \(0.40 \pm 0.01\). The rigid flippers demonstrated a similar pattern, with the 'Adaptive Gait Change' configuration yielding the most significant displacement of \(0.67 \pm 0.01\) BL/cycle, outperforming the 'Diagonal Gait' at \(0.56 \pm 0.02\) and the 'All Together Gait' at \(0.52 \pm 0.01\).

These findings verify the hypothesis that an adaptive gait mechanism, which intelligently alters locomotion strategies in real time, can optimize a robot's movement across complex and variable terrains.

\section*{Discussion}

Our study examines how flipper morphology and gait patterns influence a bio-inspired flippered quadruped robot's ability to adapt to diverse terrains, taking inspiration from the locomotion of sea turtle hatchlings. The study highlights the complex interplay between mechanical design and environmental interaction necessary for proficient movement across various complex surfaces, including granular media, flat surfaces, and inclines.

Our robotic design, characterized by its low cost, robust design, and autonomous navigation capabilities, successfully emulates the versatile locomotion of sea turtle hatchlings (see movie S11). The biomimetic approach extends beyond mere replication of movement; it capitalizes on these creatures' inherent adaptability to navigate real-world environments effectively. By integrating sensors and adaptive algorithms, the robot demonstrates significant potential for practical applications in scenarios that mimic the challenges faced by sea turtles in their natural habitats.

Through extensive testing, we have gathered valuable data pointing to strategic combinations of flipper configurations and gaits optimized for specific environmental contexts. These findings underscore the importance of flexibility in design to accommodate the unpredictable nature of real-world terrains and set the stage for future developments in autonomous robotic systems capable of sophisticated terrain adaptation.

Our study in dry sand revealed that using all flippers, soft or rigid, regardless of gait, enables effective movement. The success of front-only rigid flippers in an all-together gait suggests that certain flipper stiffnesses could enhance locomotion on such terrain with reduced flippers. Traversing rocky terrains with their inherent irregularities and obstacles highlighted the advantages of the rigid flippers' stability and support, especially when combined with an all-together gait pattern. This configuration helped the robot maintain balance and effectively transfer force to navigate the uneven ground. Wet sand posed a different challenge; the cohesive nature of the moist substrate required a gait and morphology that prevented sinking while maintaining traction. Soft flippers in a diagonal gait emerged as superior, supporting the hypothesis that softer materials are better suited for navigating surfaces where grip and the ability to mold around uneven surfaces (malleability) are essential. The experiments on sandy inclines stressed the importance of rear flipper engagement, with full flipper configurations showing far better performance in climbing steep terrains than front-only setups. The foam stairs tests further validated this, as full flipper configurations were crucial for ascending the vertical steps, where each flipper's push-and-pull action is vital for upward propulsion.

Across all these terrains, the robot's ability to maintain its trajectory and autonomously adapt its gait based on terrain recognition proved essential. Implementing an adaptive gait change mechanism allowed the robot to switch between the most effective gaits for the given surface, which was particularly evident in the multi-terrain displacement rate experiments. The robotic system demonstrated proficiency in designed terrains; however, our current approach has some limitations. While controlled environments allowed for precise measurements and repeatable conditions, they did not entirely replicate real-world terrains' unpredictable and varied nature. This limitation suggests that the robot's performance in a more complex natural setting might differ, necessitating further testing in diverse, uncontrolled environments. Future iterations of this research will incorporate a broader scope of terrain types under more variable and less controlled conditions.

Reliance on color sensors for terrain recognition poses another limitation, potentially leading to inaccuracies in less controlled lighting conditions or visually uniform but texturally diverse surfaces. Augmenting the robot's perception system with other sensory inputs, such as cameras, LIDAR, radar, or even tactile sensors, could address these challenges, enabling the robot to discern and navigate more complex environments. Moreover, our approach to flipper stiffness alteration—though effective—lacks the real-time adaptive capabilities that would allow for instantaneous morphological adjustments. A robotic system capable of actively modulating its structural rigidity similar to the robot given in previous study \cite{baines_toward_2019,baines_multi-environment_2022} in response to sensory feedback would mark a significant advancement, aligning more closely with the versatility inherent in biological organisms.

We observe that while adaptive gait changes generally enhance locomotive efficiency, the superiority of soft flippers on most terrains suggests that material compliance is often more advantageous. This aligns with the concept of ''embodied intelligence`` found in soft robotics \cite{coyle_bio-inspired_2018}, where the inherent properties of materials confer an ability to navigate and interact with complex environments more intuitively. The exceptions, noted in certain contexts with rigid flippers, point towards a nuanced balance where the structural support provided by rigidity is sometimes necessary, particularly on hard, flat surfaces. The promise shown by the adaptive gait mechanism leads us to consider machine learning algorithms for predictive and reactive gait modulation, potentially allowing the robot to learn from its environment and further refine its locomotive strategies. By integrating a broader array of sensory data and historical performance metrics, the robot could dynamically adjust not only its gait but also predict the most suitable morphological configuration for upcoming terrain changes.

Future work would also explore incorporating aquatic locomotion capabilities, reflecting the natural proficiency of sea turtles in both land and water. The transition between terrestrial and aquatic environments remains a relatively understudied aspect in robotics, one that could greatly expand the operational range of bio-inspired designs \cite{fish_advantages_2020,ijspeert_swimming_2007,corucci_evolving_2018,wang_bio-inspired_2009,baines_amphibious_2021}. Additionally, considering that our prototype emulates a sea turtle hatchling, investigating the movement of adult sea turtles and other aquatic animals with flippers could provide further valuable insights, expanding the scope and applicability of bio-inspired robotic systems in complex environments.

Overall, while this study's results are promising, they underline the need for ongoing development to overcome the current limitations and ensure that the robotic system can reliably operate in more complex and less predictable environments. The work of researchers like Ijspeert \cite{ijspeert_biorobotics_2014} and Melo et al. \cite{melo_animal_2023} provide further insights into the potential of exploring efficient and agile locomotion through biorobotics. Our future work will continue to build on this current study, improving our robot's design and functionality towards practical applicability. We would also explore considerations around the scalability of the design, which will be crucial for ensuring the robot's functionality across different sizes and applications. 

In conclusion, the insights gained from this study contribute significantly to the field of bio-inspired robotics. Our results support the notion that while a single gait or morphology might perform well in specific conditions, adapting to the changing demands of different terrains is crucial for creating versatile robotic systems. By embodying the locomotive strategies of sea turtles, we have taken a substantial step forward in developing robotic systems capable of navigating challenging terrains. The potential applications of such technology are vast, ranging from environmental monitoring to search and rescue missions. Continued advancements in this field hold great potential to unlock new possibilities for robotic interaction with the world's most demanding environments.

\section*{Materials and Methods}

\subsection*{Mechanical Design and Fabrication}\label{mechdesign}

The robot body (Fig. \ref{fig:fullimage}, movie S1) is designed to capture the morphological attributes of a baby sea turtle. The oval-shaped body frame, measuring 12.5 cm in length, was designed using Solidworks software and later 3D printed with a Stratasys F170 printer using Acrylonitrile Butadiene Styrene (ABS) material. The body houses the electronic components.

The robot is equipped with four independently servo-actuated flippers; the flexible flippers were constructed using Dragon Skin™ 20 silicone rubber of 20A shore hardness. The rubber was mixed in a 1A:1B volume ratio and formed in a mold (Fig. S3). Larger fore-flippers (length = 12.5 cm) serve as the primary propulsion mechanism and facilitate the scooping and pushing motions essential for effective movement on various terrains. The smaller hind flippers, measuring 3.5 cm in length, aid in steering and provide additional stability and maneuverability. This anatomical differentiation between fore and hind flippers mimics the functional biomechanics of sea turtles, which use their larger front limbs for powerful strokes in water or to navigate on land and their smaller rear limbs for directional control and balance \cite{renous_locomotion_2000,avens_responses_2003}. The use of Dragon Skin™ 20 silicone rubber, renowned for its durability and flexibility, ensures that the flippers can withstand the stresses of operation across different environments. The chosen shore hardness of 20A allows the flippers to maintain a balance between firmness and flexibility. The rigid flippers were constructed from ABS material and are the same size as the flexible flippers. They were coated with rubber film (Plasti Dip) to reduce the impact of friction.

\subsection*{Electronics and Control Mechanism}\label{electronics} 
The electronics and control mechanisms of our robot are designed to provide the required functionality for efficient and adaptive locomotion across diverse environments. The robot has six Robotis Dynamixel XL330-M288-T motors (0.42 N.m) for flipper actuation, a Robotis OpenRB-150 microcontroller for onboard control, an array of sensors to optimize its navigation and performance assessment, and a 3.7V, 2000 mAh lithium polymer battery (see Fig. \ref{fig:fullimage}D). The motors are well-enclosed to prevent sand damage to the gears and control the precise movements of the flippers through the angular actuation of the rotors. The sensor array includes an Inertial Measurement Unit (IMU), specifically the Adafruit LSM6DS3TR-C 6-DoF Accel + Gyro IMU, to ensure that the robot maintains a consistent travel direction, thereby improving the repeatability of experiments, a power sensor, the Adafruit INA260, measures the robot's power consumption, which is instrumental in calculating the cost of transport. In addition, a color sensor (Adafruit TCS34725) has been attached to the front part of the robot to recognize different terrains, facilitating adaptive gait transitions essential for efficient locomotion across diverse environmental conditions.

\subsection*{Robot Kinematics}
The kinematic model of our quadruped robot, inspired by the locomotive mechanisms of sea turtles, is fundamental to understanding its movement capabilities and efficiency across different terrains. This model forms the basis for analyzing the robot's movements, leading to more effective design iterations and control strategies. This model only considers the rigid flipper configuration to simplify the kinematic equations, and the kinematic diagram is shown in Fig. S4.

\textbf{Kinematic Configuration.} Our robot's design mimics the flipper structure of a sea turtle, featuring two primary joints in each front flipper, akin to a simplified shoulder and elbow joint system. Each flipper's kinematic chain consists of two revolute joints, allowing pitch and yaw movements (Fig. \ref{fig:fullimage}A, Fig. S1A-B).

\textbf{Joint 1 - Pitch Motion.} The first joint, analogous to a shoulder joint, facilitates the up-and-down (pitch) motion. This joint is directly attached to the robot's body and rotates within -90 to 90 degrees. At 0 degrees, it is aligned parallel to the robot's main body.

\textbf{Joint 2 - Yaw Motion.} The second joint, like an elbow, provides the left-and-right (yaw) motion. It is connected to the first joint via a short link, which emulates the sea turtle's flipper bone structure. This joint also has a motion range of -90 to 90 degrees, and at 0 degrees, it is perpendicular to the robot's body, allowing for effective maneuvering in granular media.

\textbf{Forward Kinematics.}
The forward kinematics of each flipper are derived by considering the sequential transformations from the robot's base frame to the flipper's endpoint. Let \( L_1 \) and \( L_2 \) denote the lengths of the first and second links of the flipper, corresponding to the distance between the two joints and the length of the flipper, respectively.

\textbf{Transformation Matrices.}
Given the joint angles \( \theta_1 \) and \( \theta_2 \) for the pitch and yaw motions, the transformation matrices can be expressed as follows:

\begin{itemize}
  \item Transformation from base to Joint 1 (\( T_{1} \)):
  \[ T_{1} = \begin{bmatrix}
  \cos(\theta_1) & 0 & \sin(\theta_1) & 0 \\
  0 & 1 & 0 & 0 \\
  -\sin(\theta_1) & 0 & \cos(\theta_1) & 0 \\
  0 & 0 & 0 & 1
  \end{bmatrix} \]

  \item Transformation from Joint 1 to Joint 2 (\( T_{2} \)):
  \[ T_{2} = \begin{bmatrix}
  \cos(\theta_2) & -\sin(\theta_2) & 0 & L_1\cos(\theta_2) \\
  \sin(\theta_2) & \cos(\theta_2) & 0 & L_1\sin(\theta_2) \\
  0 & 0 & 1 & 0 \\
  0 & 0 & 0 & 1
  \end{bmatrix} \]

  \item Transformation from Joint 2 to the flipper's endpoint (\( T_{3} \)):
  \[ T_{3} = \begin{bmatrix}
  1 & 0 & 0 & L_2 \\
  0 & 1 & 0 & 0 \\
  0 & 0 & 1 & 0 \\
  0 & 0 & 0 & 1
  \end{bmatrix} \]
\end{itemize}

\textbf{Endpoint Position.}
The position of the flipper's endpoint relative to the robot's base frame is obtained by multiplying these transformation matrices:

\[ P = T_{1} \cdot T_{2} \cdot T_{3} \cdot \begin{bmatrix} 0 \\ 0 \\ 0 \\ 1 \end{bmatrix} \]

This results in a matrix where the first three elements of the last column represent the x, y, and z coordinates of the flipper's endpoint in the base frame.



\subsection*{Cost of Transport}

Cost of Transport (CoT) is critical in assessing locomotion efficiency, allowing meaningful comparisons across different robots and organisms \cite{von1950price,tucker_energetic_1975,gregorio_design_1997}. It is a nondimensional metric defined as the ratio of power input ($P_{in}$) to the product of the robot's mass (m), acceleration due to gravity ($g$), and forward speed ($v$), CoT serves as a reliable measure of energy expenditure for unit mass over a unit distance ($d$). The formula for CoT is given as 
\begin{equation}
COT = \frac{P_{in}}{mgv} = \frac{E_{total}}{mgd}
\end{equation}

Where ${P_{in}}$ is calculated as the sum of the power drawn from each motor on the shoulder, elbow, and hip joint, given by the formula:

\begin{equation}
P_{in} = \sum_{n=1}^{6} I_{n}V
\end{equation} 
$I_{n}$ is the average current drawn from the nth motor on each joint (measured in amperes) during the gait, and $V$ is the constant voltage supplied to the robot (3.7V). The CoT was calculated for three trials of each gait. Considering the different terrains, the gait with the lowest CoT indicated the most energy-efficient configuration.





\section*{Acknowledgments}
We would like to thank the Naughton Undergraduate Fellowship program for supporting John McElroy. We would also like to express our appreciation to the members of Notre Dame MiNiRo-Lab for their invaluable contributions and insightful discussions.

\section*{Author Contributions}
N.C., J.M., and Y.O.A designed research; N.C. and J.M., conceived the setups and performed experiments; N.C., J.M., and Y.O.A. analyzed data; N.C. and Y.O.A. wrote the paper.





\clearpage


\end{document}